\pdfoutput=1

\documentclass[11pt]{article}
\usepackage[final]{acl}

\usepackage{times}
\usepackage{float}
\usepackage{latexsym}
\usepackage{graphicx}
\usepackage{subfigure}
\usepackage[T1]{fontenc}
\usepackage{amsmath} 
\usepackage{amsthm}
\usepackage{bm}

\usepackage[utf8]{inputenc}

\usepackage{microtype}
\usepackage{multirow}
\usepackage{inconsolata}
\usepackage{amssymb}
\usepackage{comment}
\usepackage{tikz}
\usepackage{enumitem}

\def\1{\bm{1}}

\def\ro{{\textnormal{o}}}
\def\rp{{\textnormal{p}}}

\def\rs{{\textnormal{s}}}

\def\rva{{\mathbf{a}}}

\def\rvh{{\mathbf{h}}}

\def\rvo{{\mathbf{o}}}

\def\rvw{{\mathbf{w}}}

\def\rmH{{\mathbf{H}}}

\def\rmR{{\mathbf{R}}}

\def\rmW{{\mathbf{W}}}

\def\ermA{{\textnormal{A}}}
\def\ermB{{\textnormal{B}}}

\def\vh{{\bm{h}}}

\def\vq{{\bm{q}}}
\def\vr{{\bm{r}}}

\def\vx{{\bm{x}}}
\def\vy{{\bm{y}}}
\def\vz{{\bm{z}}}

\def\mW{{\bm{W}}}

\DeclareMathAlphabet{\mathsfit}{\encodingdefault}{\sfdefault}{m}{sl}
\SetMathAlphabet{\mathsfit}{bold}{\encodingdefault}{\sfdefault}{bx}{n}
\newcommand{\tens}[1]{\bm{\mathsfit{#1}}}

\def\tK{{\tens{K}}}

\def\tP{{\tens{P}}}
\def\tQ{{\tens{Q}}}

\def\tV{{\tens{V}}}

\def\tX{{\tens{X}}}
\def\tY{{\tens{Y}}}

\def\gD{{\mathcal{D}}}

\def\gL{{\mathcal{L}}}

\def\gS{{\mathcal{S}}}
\def\gT{{\mathcal{T}}}

\def\sD{{\mathbb{D}}}

\def\sR{{\mathbb{R}}}

\newcommand{\softmax}{\mathrm{softmax}}

\DeclareMathOperator{\LLM}{LLM}
\DeclareMathOperator{\PMA}{PMA}
\DeclareMathOperator{\LN}{LN}
\DeclareMathOperator{\MHA}{MHA}
\DeclareMathOperator{\FFN}{FFN}

\DeclareMathOperator{\corr}{corr}
\DeclareMathOperator{\similarity}{sim}

\newcommand*\circled[1]{\tikz[baseline=(char.base)]{
            \node[shape=circle,draw,inner sep=2pt,scale=0.6] (char) {#1};}}

\newcommand{\astmark}[1][*]{\textsuperscript{#1}}
\newcommand{\customdagger}{\textsuperscript{\textdagger}}
\title{D2LLM: Decomposed and Distilled Large Language Models\\for Semantic Search}

\author{Zihan Liao\thanks{Equal contribution. This work was done when Zihan Liao was a research intern at Ant Group.} \\ East China Normal University \\ \texttt{52215901015@stu.ecnu.edu.cn} \And
        Hang Yu\astmark \\ Ant Group \\ \texttt{hyu.hugo@antgroup.com} \AND
        Jianguo Li\thanks{Corresponding authors.} \\ Ant Group \\ \texttt{lijg.zero@antgroup.com} \And
        Jun Wang\customdagger \\ East China Normal University \\ \texttt{wongjun@gmail.com} \And
        Wei Zhang\customdagger \\ East China Normal University \\ \texttt{zhangwei.thu2011@gmail.com}
        }
\begin{document}
\maketitle
\begin{abstract}
The key challenge in semantic search is to create models that are both accurate and efficient in pinpointing relevant sentences for queries. While BERT-style bi-encoders excel in efficiency with pre-computed embeddings, they often miss subtle nuances in search tasks. Conversely, GPT-style LLMs with cross-encoder designs capture these nuances but are computationally intensive, hindering real-time applications. In this paper, we present D2LLMs—Decomposed and Distilled LLMs for semantic search—that combines the best of both worlds. We decompose a cross-encoder into an efficient bi-encoder integrated with Pooling by Multihead Attention and an Interaction Emulation Module, achieving nuanced understanding and pre-computability. Knowledge from the LLM is distilled into this model using contrastive, rank, and feature imitation techniques. Our experiments show that D2LLM surpasses five leading baselines in terms of all metrics across three tasks, particularly improving NLI task performance by at least 6.45\%. The source code is available at \url{https://github.com/codefuse-ai/D2LLM}.
\end{abstract}

\section{Introduction}
\label{sec:intro}

Semantic search has become an integral part of natural language processing, tasked with sifting through extensive texts to find passages that best match a user's query based on underlying semantic links. 
It transcends the non-semantic techniques, such as TF-IDF and BM25, by resolving lexical mismatches and enabling more precise text matching.
As a result, semantic search has significant impacts across various fields, including information retrieval~\cite{zhu2023large}, question answering~\cite{allam2012question}, dialogue systems~\cite{chen2017survey},  item recommendation~\cite{hu2020graph}, fact checking~\cite{thorne2018fever}, and retrieval-augmented generation~\cite{gao2024retrievalaugmented}.

The major challenge of semantic search lies in devising a model that is both \textbf{accurate} and \textbf{efficient} in pinpointing the most relevant passages for any given query. The current go-to models, particularly the compact BERT-style bi-encoders or dual encoders~\cite{wang2022text,xiao2023c,li2023towards}, independently convert queries and passages into vectors and judge their relevance through measures such as cosine similarity. This process is praised for \textbf{efficiency}—enabling pre-computation and on-the-fly querying of passage vectors. However, this streamlined method comes at an \textbf{accuracy} cost. Within the rigidity of the bi-encoder's similarity space, subtle but critical nuances may be lost, such as when differentiating between symmetric search tasks (e.g., finding similar questions to "What are the symptoms of the flu?") and asymmetric search tasks (e.g., matching that same query to a comprehensive answer detailing symptoms). The bi-encoders' constrained interaction mode limits their comprehension of the distinct informational roles that queries and passages play. Additionally, bi-encoders are bound to a laborious and multi-stage training process, starting with pretraining on massive datasets of weakly-supervised text pairs and ending with finetuning on diverse and extensive labeled datasets~\cite{wang2023improving}. This process is heavily data-intensive and usually limited by the variety of data available. Moreover, the small size of bi-encoders often means they excel within their training domain but fall short when generalizing to new, unseen domains~\cite{rosa2022defense,rosa2022no,su2022one}.

On the flip side, GPT-style Large language models (LLMs) with cross-encoder designs overcome these limitations by jointly processing queries and passages, thereby forming a single, interactive input. This method enables a granular understanding of textual relationships, as it involves concatenating the query and passage, with directive prompts such as "Are these questions similar?" or "Does this passage answer the question?" to guide the model through both symmetric and asymmetric search tasks. Updated LLMs arrive pre-loaded with a broad spectrum of world knowledge~\cite{hu2023language}, eliminating the need for domain-specific pretraining and facilitating rapid adaptation. The remarkable zero-shot learning ability ~\cite{wei2021finetuned,kojima2022large} ensures their robust performance even for novel domains. However, this \textbf{accurate} analysis incurs a toll on computational \textbf{efficiency}; it precludes the caching of passage vectors and necessitates fresh inference for each new query-passage pairing, which can hinder the practicality of cross-encoder LLMs in situations demanding real-time, voluminous processing.

In this paper, we seek to bring the best of both worlds together with the introduction of \textbf{D2LLM}s, which stands for \textbf{Decomposed and Distilled LLM}s for semantic search. Our proposed framework begins by \textbf{decomposing} an LLM-based cross-encoder into a bi-encoder coupled with an Interaction Emulation Module (IEM). The bi-encoder, equipped with Pooling by Multihead Attention (PMA) of token embeddings resulting from a pretrained LLM, efficiently generates separate embeddings for queries and passages, allowing passage vectors to be pre-stored while ensuring the model's adaptability. The IEM goes further, intricately mapping the relationships between queries and passages. It features dedicated branches for handling symmetric and asymmetric search tasks. We then \textbf{distill} the high-level knowledge from the original LLM-based cross-encoder (the teacher) into our decomposed model (the student) through a series of teacher-guided methodologies, including contrastive imitation, rank imitation, and feature imitation. 
Our contributions can be summarized as:
\begin{itemize}[leftmargin=*, topsep=1pt,itemsep=1pt,partopsep=1pt,parsep=1pt]
    \item We introduce D2LLM, a new semantic search solution that combines the speed of bi-encoders with the accuracy of LLM-based cross-encoders. This method breaks down the complex cross-encoder into a more manageable student model comprising a bi-encoder, a PMA, and an IEM.
    \item We transfer the teacher's linguistic expertise to the student through a comprehensive knowledge distillation strategy, encompassing contrastive imitation, rank imitation, and feature imitation techniques.
    \item Our empirical results reveal that D2LLM outperforms five leading methods in three benchmark tasks, with a particularly notable 14.39\% improvement over the second-best LLaRA and a 6.45\% lead over the heavily finetuned benchmark BGE model in the NLI task.
\end{itemize}

\section{Related Works}
\paragraph{Classical Models}
Classical semantic search techniques leveraging more compact language models can generally be categorized into bi-encoders, cross-encoders, and bi-encoders distilled from cross-encoders. The first approach, bi-encoders, often finetunes BERT-style models through contrastive learning, with a focus on negative sample mining, as exemplified by SBERT~\cite{reimers2019sentence}, ANCE~\cite{xiong2020approximate}, DPR~\cite{karpukhin2020dense}, SimCSE~\cite{gao2021simcse}, ME-BERT~\cite{luan2021sparse}, RocketQA~\cite{qu2021rocketqa}, ADORE~\cite{zhan2021optimizing}, and DiffCSE~\cite{chuang2022diffcse}. However, finetuning alone does not always ensure adaptability and generalization, leading to the integration of self or weakly supervised pretraining as seen in ICT~\cite{lee2019latent}, Condenser~\cite{gao2021your}, Cocondenser~\cite{gao2022unsupervised}, Contriever~\cite{izacard2022unsupervised}, OpenAI Embeddings~\cite{neelakantan2022text}, E5~\cite{wang2022text}, GTE~\cite{li2023towards}, and BGE~\cite{xiao2023c}. Despite this, these models still face challenges in capturing complex query-passage relationships. While multiview embeddings have been suggested~\cite{zhang2022multi}, cross-encoders~\cite{rosa2022defense,rosa2022no}—our second category—address this more effectively but are not well-suited for real-time use due to the computational cost of recomputing passage representations. The third category, most related to our proposed D2LLM, aims to distill the effectiveness of cross-encoders into bi-encoders. Previous work in this area has primarily used straightforward distillation strategies, such as pseudo-labeling~\cite{qu2021rocketqa,ren2021pair,izacard2021distilling} or KL divergence loss~\cite{yang2020retriever}, to align the student model with the teacher model. More advanced techniques like AR2~\cite{zhang2022adversarial} and RocketQAv2~\cite{ren2021rocketqav2} have attempted joint optimization of student and teacher models using adversarial training and dynamic distillation. However, these approaches face challenges when applied to LLMs: scalability issues due to the co-training of two models, loss functions that may not capture the full breadth of knowledge in LLMs, and the failure to adequately model the detailed interactions between queries and passages.

Our D2LLM framework overcomes these challenges by focusing on refining the student model through distillation with a static teacher model, which can be pre-finetuned for specific tasks. We enhance the distillation process by employing a combination of contrastive, rank, and feature imitation losses. Crucially, we integrate an Interaction Emulation Module into the student model to better understand and replicate the nuanced interplay between queries and passages, thus solving the problems of previous distillation approaches.

\paragraph{LLMs}
Classical semantic search methods often suffer from limited size, leading them to excel in one specific task like text similarity, classification, or information retrieval, as with SimCSE~\cite{gao2021simcse} and Contriever~\cite{izacard2022unsupervised}. Yet, larger models and improved pretraining techniques have been shown to significantly boost both the accuracy and applicability of PLM-based dense retrieval~\cite{izacard2022unsupervised,wang2022text,xiao2023c}. This has led to the application of LLMs to semantic search, though this field remains underexplored. Current approaches to transforming LLMs into bi-encoders include finetuning only methods like SGPT~\cite{muennighoff2022sgpt}, RepLLaMa~\cite{ma2023fine}, Udever~\cite{zhang2023language}, and LLaRA~\cite{li2023making}, as well as those combining continued pretraining with finetuning, such as CPT~\cite{neelakantan2022text} and GTR~\cite{ni2022large}. Typically, these methods introduce a special token at the end of the passage, trained with contrastive loss to emulate the role of BERT's [CLS] token. However, this adaptation deviates from the original training objective of LLMs, which is next-token prediction, potentially underutilizing LLMs' capabilities and resulting in suboptimal performance compared to smaller, specialized bidirectional encoders like GTE~\cite{li2023towards} and BGE~\cite{xiao2023c}. 

The proposed D2LLM tackles this issue by decomposing the teacher LLM into a student bi-encoder combined with an Interaction Emulation Module that more faithfully mirrors the LLM's handling of query-passage interactions. Through a detailed distillation process, D2LLM effectively transfers knowledge from the teacher LLM, producing a student model that excels beyond the aforementioned models finetuned on the same datasets.

\begin{figure*}
    \centering
    \includegraphics[width = \linewidth]{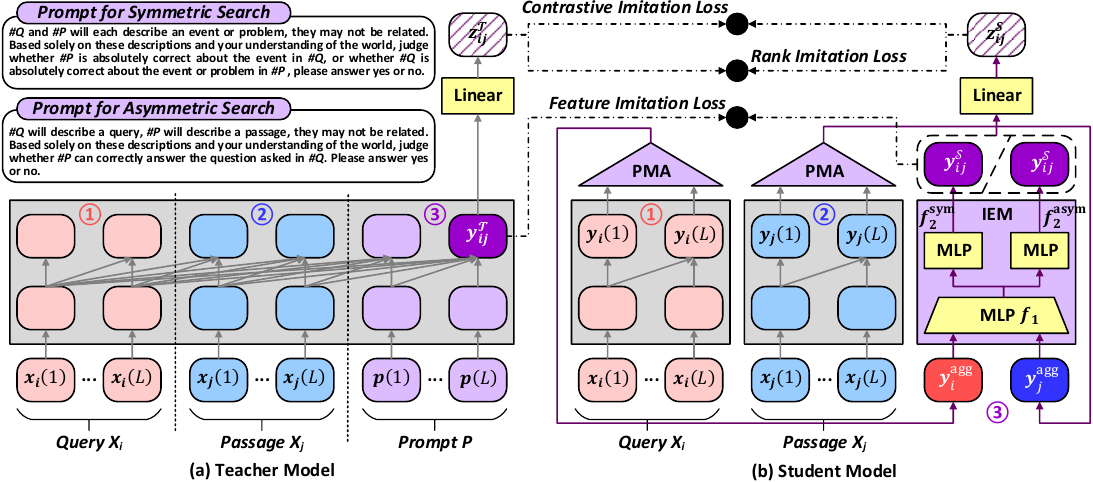}
    \vspace{-4ex}
    \caption{Architecture of D2LLM: The teacher model is decomposed into three segments corresponding to the input of the \protect\circled{1} query, \protect\circled{2} passage, and \protect\circled{3} prompt, represented in light red, light blue, and light purple. Its output, represented by a dark purple square, is the classification token embedding, which, after a linear layer, yields logits. The student model maintains the query and passage components (\protect\circled{1} and \protect\circled{2}) from the teacher but adds \protect\circled{3} the PMA and IEM to capture the interplay between the query and passage, as well as their combined interaction with the prompt. It also outputs classification token embeddings respectively for symmetric and asymmetric search and then derives logits via a linear layer.}
    \label{fig:D2LLM}
    \vspace{-3ex}
\end{figure*}

\section{D2LLM}
\label{sec:method}
The architecture of D2LLM is depicted in Figure~\ref{fig:D2LLM}. Here, the teacher model operates as an LLM-based cross-encoder, adeptly handling joint query-passage inputs to leverage the LLM's robust semantic comprehension for more accurate outcomes. On the other hand, the student model features a bi-encoder for pre-computing passage vectors to ensure operational efficiency, complemented by an Interaction Emulation Module that considers the complex query-passage interplay. Through an elaborate knowledge distillation procedure, we aim to cultivate a student model that emulates the teacher's accuracy while retaining efficiency. Subsequently, we detail the process of decomposing the teacher model to assemble the student model and elucidate the knowledge distillation methodology.

\subsection{Decomposition}

Before delving into the construction of the student model, we first introduce how an LLM plays the role of the teacher. 

\subsubsection{Teacher Model $\gT$}
The teacher model aims to accurately determine whether a query $\tX_i$ and a passage $\tX_j$ are a compatible match. We utilize a cross-encoder architecture for this purpose since it leverages the LLM's strength in making informed decisions by synthesizing different parts of the combined inputs. To tap into the LLM's zero-shot learning capabilities, we employ prompt engineering, crafting specific prompts $\tP$ that guide the LLM in analyzing the query-passage pairs. As indicated in the upper left panel of Figure~\ref{fig:D2LLM}, we design prompts $\tP$ for symmetric and asymmetric searches ($\tP^\text{sym}$ and $\tP^\text{asym}$). The chosen prompt $\tP\in\{\tP^\text{sym}, \tP^\text{asym}\}$ is then concatenated with the query-passage pair $(\tX_i, \tX_j)$, prompting the LLM to generate a ``yes'' or ``no'' response. This process is represented in Figure~\ref{fig:D2LLM}(a), where the LLM generates an embedding for the response (i.e., the dark purple square): 
\begin{align} \label{eq:teacher_token}
    \vy_{ij}^\gT = \LLM(\tX_i, \tX_j, \tP).
\end{align}
The embedding $\vy_{ij}^\gT$ functions as a classification token that labels the query-passage pair $(\tX_i, \tX_j)$ as related or not, with the prompt $\tP$ aiding in adapting to different search task types (symmetric versus asymmetric). To determine the probability of ``yes'' or ``no'', we extract the corresponding rows from the weight matrix $\mW^\gT$ in the last layer of the original LLM, compute the logits $\vz_{ij}^\gT$, and finally apply a softmax function to calculate the probability of ``yes'' (i.e., score) $s_{ij}^\gT$, that is,
\begin{align}
    \vz_{ij}^\gT &= \mW^\gT[\text{``yes'',``no''}]\vy_{ij}^\gT, \label{eq:teach_logits} \\
    s_{ij}^\gT &= \softmax(\vz_{ij}^\gT). \label{eq:teach_score}
\end{align}
In practice, it is more convenient to compute the logits for all tokens within the LLM and subsequently extract the probabilities $s_{ij}^\gT$ for "yes" and "no" by inputting the relevant logits into the softmax function. Beyond prompt engineering, the teacher model can also be finetuned for improved performance.

\subsubsection{Student Model $\gS$}

As shown in Figure~\ref{fig:D2LLM}(a), the teacher model can be decomposed into three components processing the input of \protect\circled{1} the query, \protect\circled{2} the passage, and \protect\circled{3} the prompt, respectively. The third component goes beyond handling the prompt; it integrates and examines the interactions between the query and passage, as well as their collective interplay with the prompt, to ultimately determine their match. Mirroring the teacher, the student model retains the same structure for handling \protect\circled{1} the query and \protect\circled{2} the passage. Innovatively, for \protect\circled{3}, it assimilates the prompt information and the query-passage-prompt interactions via the Pooling by Multihead Attention (PMA) and the Interaction Emulation Module (IEM).

\paragraph{PMA:}The PMA module \cite{lee2019set} synthesizes information from tokens within the query $\tX_i$ and the passage $\tX_j$, producing a distinct embedding vector for each.  For a query of length $L$, represented as $\tX_i = [\vx_i(1),\ldots,\vx_i(L)]$ with $\vx_i(k)$ being the $k$-th token, and corresponding hidden states $\tY_i = [\vy_i(1),\ldots,\vy_i(L)]$ from the LLM's last layer, PMA aggregates token information as:
\begin{align}
    \vy_i^\text{agg} &= \PMA_\vq(\tY_i) = \LN(\vh + \FFN(\vh)), \\
    \vh &= \LN(\MHA(\vq, \tY_i, \tY_i) + \vq),
\end{align}
where $\LN$, $\FFN$, and $\MHA(\tQ, \tK, \tV)$ respectively denote layer normalization, feedforward network, and multihead attention with query $\tQ$, key $\tK$, and value $\tV$. The PMA's query $\vq$ is a learnable vector that functions as an anchor, extracting information from the $L$ tokens based on their similarity to the query $\vq$ for semantic search. This attention-based pooling offers greater flexibility than traditional mean/min/max pooling by allowing dynamic weight adjustments, as opposed to fixed weights for all instances of $\tX_i$ and $\tX_j$. 

Note that, unlike BERT-style models which typically use a special token [CLS] for sentence embedding, GPT-style models lack an equivalent mechanism. Alternative methods for GPT-style models, like CPT~\cite{neelakantan2022text} and LLaRA~\cite{li2023making}, have tried using the last token or the [EOS] token as a substitute for the [CLS] token. However, such usage diverges from the LLMs' intended next-token prediction function, often to their detriment. Alternatively, SGPT~\cite{muennighoff2022sgpt} introduces position-weighted pooling, but may impose undue constraints by presuming that token relevance is a function of their positions. PMA is favored as it does not conflict with LLMs' inherent nature and its learnability ensures adaptability.

\paragraph{IEM:}After the PMA generates individual vectors for the query and passage, the IEM implicitly encodes the prompt (be it symmetric or asymmetric) and captures the query-passage interaction. As depicted in Figure~\ref{fig:D2LLM}(b)'s right panel, we concatenate the query and passage embeddings and input them into a Multi-Layer Perceptron (MLP) designed with two branches to handle the respective prompt nuances, which can be expressed as:
\begin{align}
    \vy_{ij}^\gS = f_2(f_1([\vy_i^\text{agg},\vy_j^\text{agg}])),
\end{align}
where both $f_1$ and $f_2$ are MLPs. $f_1$ extracts elementary features from the combined embeddings, while $f_2 \in \{f_2^\text{sym}, f_2^\text{asym}\}$ is tailored to the two branches, processing symmetric and asymmetric searches. After deriving $\vy_{ij}^\gS$, the student model computes its logits $z_{ij}^\gS$ and score $s_{ij}^\gS$ in a manner akin to the teacher model (cf. Eqs.~\eqref{eq:teach_logits}-\eqref{eq:teach_score}), but with a learnable linear layer $\mW^\gS$. We underscore that the MLP operates as a flexible similarity metric, enhancing the description of the query-passage relationship beyond the commonly used cosine similarity in bi-encoders. To maintain efficiency, lightweight MLPs are utilized. Please refer to Appendix~\ref{app:iem} for more discussion on IEM.


\subsection{Distillation}

Knowledge distillation aims to impart the teacher model's capabilities to the student model. To accomplish this, we focus on three specific training objectives: contrastive, rank, and feature imitation.

\subsubsection{Contrastive Imitation}
\label{ssec:ci}
For a given query $\tX_i$, we curate a set of positive samples $\sD^+$ (relevant passages) and negative samples $\sD^-$ (irrelevant passages). The negative sample set includes in-batch negatives and hard negatives, the latter sourced from the top-$k$ results using BM25 and bi-encoder evaluations~\cite{ma2023fine,li2023making,li2023towards}, thus forming $\sD^- = \sD^-_I \cup \sD^-_H$. Note that a few hard negatives may potentially be latent positives, but our Contrastive Imitation can address this circumstance robustly, which will be discussed later. The resulting Contrastive Imitation (CI) loss is:
\begin{align}
    \gL^\text{CI} = -\frac{1}{|\sD^+|}\sum_{j\in\sD^+}\log \dfrac{\exp(s_{ij}^\gT z_{ij}^\gS/\tau)}{\displaystyle\sum_{k\in\sD^-}\exp((1-s_{ik}^\gT) z_{ik}^\gS/\tau)}, \notag
\end{align}
where $\tau$ is the temperature parameter, $s_{ij}^\gT$ is the teacher's probability score for a "yes" between pairs $(i,j)$, and $z_{ij}^\gS$ is the student's corresponding logit (unnormalized probability). The CI loss diverges from traditional contrastive loss by leveraging the teacher's scores $s^\gT$ to account for varying relevance among samples, assigning higher weights to easy negatives than hard ones. Even if true positives incorrectly appear in $\sD^-$, the CI loss remains unaffected, ensuring a more robust training environment than standard contrastive loss. It also emphasizes positive samples with higher teacher scores, indicating their criticality. 

\subsubsection{Rank Imitation}
\label{ssec:ri}
While Contrastive Imitation (CI) loss effectively handles true positives and easy negatives, it does not adequately address the gradations among samples. This is where Rank Imitation (RI) steps in, focusing on distinguishing between positive and hard negative samples, as well as discerning easy from hard negatives, thus enabling the student to replicate the teacher's subtle ranking nuances.

To synchronize the student's and the teacher's ranking of positive and hard negative samples, we aim to maximize the Pearson correlation~\cite{huang2022knowledge} between their logits. The RI loss dedicated to this alignment is:
\begin{align}
    \gL^\text{RI}_{PH} = 1 - \corr(\vz_i^\gT, \vz_i^\gS), \label{eq:loss_ri_ph}
\end{align}
where $\vz_i^\gT = [z_{ij}^\gT]$ for $j \in \sD^+ \cup \sD^-_H$ signifying a vector of the teacher's logits for the combined set of positive and hard negative samples, and likewise for $\vz_i^\gS$. We intentionally exclude in-batch negatives from this measure as they are generally easy negatives and lack the comparative relevance needed for meaningful ranking against the query $\tX_i$.

On the other hand, differentiating between hard and easy negatives is critical since hard negatives have some connection to the query, unlike easy negatives. To emphasize this, we introduce an additional RI loss for these two groups of samples:
\begin{align}
    \gL^\text{RI}_{HI} =&\ - \frac{1}{|\sD^-_H||\sD^-_I|}\sum_{j\in\sD^-_H}\sum_{k\in\sD^-_I} \lambda_{jk}\log(\sigma(z_{ij}^\gS \notag \\
    &\ - z_{ik}^\gS)), \label{eq:loss_ri_hi}
\end{align}
where $\lambda_{jk}$ is the rank comparison metric between a hard negative $j$ and an in-batch negative $k$ as determined by the teacher. The metric utilized is the normalized discounted cumulative gain (NDCG)~\cite{jarvelin2002cumulated}. It gives a non-zero $\lambda_{jk}$ only when $z_{ij}^\gT-z_{ik}^\gT > 0$ in the teacher model. The design of this loss ensures that only when a sample is deemed an easy negative by the teacher does the student assign it a lower score than it would to a hard negative (i.e., $z_{ij}^\gS-z_{ik}^\gS>0$). This approach allows the student to effectively differentiate between easy and hard negatives under the teacher's guidance, even when hard negatives are interspersed with in-batch negatives.

\subsubsection{Feature Imitation}
\label{ssec:fi}
CI and RI aim to align the student's output with that of the teacher, emphasizing output distillation. Working in tandem, feature distillation can also provide substantial benefits by leveraging the rich information encompassed in the teacher LLM. Directly aligning the embeddings of the classification token between the teacher and student models (i.e., $\vy_{ij}^\gT$ and $\vy_{ij}^\gS$) presents challenges due to the distinct architectures of the third component (see \protect\circled{3} in Figure~\ref{fig:D2LLM}). However, the relative relationships between embeddings for different query-passage pairs are less susceptible to such architecture variations~\cite{liu2019knowledge}. For instance, given two positive samples $\tX_j$ and $\tX_k$ as well as a negative sample $\tX_m$ for the same query $\tX_i$, $\vy_{ij}^\gT$ is often closer to $\vy_{ik}^\gT$ than $\vy_{im}^\gT$ in the feature space of the teacher, in order to produce a higher score for the pairs $(i,j)$ and $(i,k)$ than for $(i,m)$, and likewise for the student. To leverage this robustness, Feature Imitation (FI) first computes a similarity matrix for all query-passage pairs within a batch for the teacher as:
\begin{align}
    r_{ijk}^\gT = \similarity(\vy_{ij}^\gT, \vy_{ik}^\gT), \ \forall j,k\in\sD^+\cup\sD^-_H,
\end{align}
where $\similarity$ denotes cosine similarity, and repeats the process for the student to obtain $r_{ijk}^\gS$. Note that the above similarity metric is evaluated between $\vy_{ij}^\gT$ and $\vy_{ik}^\gT$, not the passage embeddings $\vy_j^\text{agg}$ and $\vy_k^\text{agg}$ alone. The goal of FI is to minimize the $\ell_2$ norm of the difference between the teacher's and student's similarity matrices for all positive and hard negative sample combinations (i.e., $\vr_i^\gT = [r_{ijk}^\gT]$ and $\vr_i^\gS = [r_{ijk}^\gS]$ for all $j$ and $k$): 
\begin{align}
    \gL^\text{FI} = \|\vr_i^\gT - \vr_i^\gS\|_2^2. \label{eq:loss_fi}
\end{align}
This approach guides the student to mimic the relational patterns in the teacher's representations, resulting in a deeper form of knowledge transfer.

\subsubsection{Overall Loss}
The collective loss is defined as a weighted sum of the above individual losses:
\begin{align}\label{eq:loss_overall}
\gL= \gL^\text{CI} + \alpha\gL^\text{RI}_{PH} + \beta\gL^\text{RI}_{HI} + \gamma\gL^\text{FI}, 
\end{align}
with the weights $\alpha$, $\beta$, and $\gamma$. 
This loss is used to train the student, including the PMA, the IEM, the linear layer $\mW^S$, and the bi-encoder (i.e., \protect\circled{1} and \protect\circled{2} in Figure~\ref{fig:D2LLM}(b)). Note that the bi-encoder is trained using parameter-efficient finetuning methods, such as LoRA~\cite{hu2021lora} and QLoRA~\cite{dettmers2023qlora}. These strategies can enhance performance without imposing a significant increase in the number of learnable parameters. Remarkably, the learnable parameters in the bi-encoder are less than 4\% of the LLM's total parameter count.

\begin{table*}[t]
\centering
\caption{Results for NLI, with the best-performing method and the second-best results marked in \textbf{bold} and \underline{underlined}, respectively.}
\label{tab:nli_results}
\vspace{-1.5ex}
\resizebox{.67\linewidth}{!}{ 
\begin{tabular}{ccccccccccc}
\hline
Dataset           &        &          & \multicolumn{4}{c}{OCNLI}         & \multicolumn{4}{c}{CMNLI}         \\ \hline
Metric     & \#Data & \#Param. & ACC    & AP     & Prec.  & Recall & ACC    & AP     & Prec.  & Recall \\ \hline
BGE-ft(326M)    & 0.3M   & 326M     & 0.5463 & 0.5689 & 0.5466 & 0.5702 & 0.6097 & 0.6656 & 0.6174 & 0.6278 \\
RocketQAv2(326M) & 0.3M   & 326M     & 0.5603 & 0.5633 & 0.5123 & 0.5723 & 0.6164 & 0.6779 & 0.5966 & 0.6905 \\
SGPT(7B)       & 0.3M   & 0.4M     & 0.5977 & 0.6165 & 0.6029 & 0.5994 & 0.6598 & 0.7259 & 0.6643 & 0.6727 \\
Udever(7B)     & 0.3M   & 89M      & 0.6412 & 0.6811 & 0.6478 & \underline{0.6698} & \underline{0.7234} & \underline{0.7819} & 0.7077 & 0.7306 \\
LLaRRA(7B)     & 0.3M   & 89M      & \underline{0.6612} & \underline{0.7115} & \underline{0.6618} & 0.6889 & 0.7150 & 0.7815 & \underline{0.7125} & \underline{0.7381} \\
\textbf{D2LLM}(7B) & 0.3M & 89M & \textbf{0.7889} & \textbf{0.8145} & \textbf{0.7736} & \textbf{0.8149} & \textbf{0.8014} & \textbf{0.8759} & \textbf{0.7960} & \textbf{0.8241} \\ 
Improv.(\%) & N/A & N/A & 19.31\% & 14.48\% & 17.30\% & 21.66\% & 10.78\% & 12.02\% & 11.72\% & 11.65\% \\ \hline
BGE(326M)        & 100M   & 326M     & 0.7266 & 0.7646 & 0.7362 & 0.7191 & 0.7675 & 0.8580 & 0.7891 & 0.7381 \\
LLM-be(7B)       & N/A    & N/A      & 0.5219 & 0.5083 & 0.5155 & 0.5955 & 0.5619 & 0.6175 & 0.5624 & 0.6762 \\
LLM-ce(7B)    & N/A    & N/A      & 0.8776 & 0.9609 & 0.8409 & 0.9493 & 0.8347 & 0.9417 & 0.8263 & 0.9303 \\ \hline
\end{tabular}}
\vspace{-1.5ex}
\end{table*}

\begin{table*}[t]
\label{sft}
\caption{Results for STS, with the best-performing method and the second-best results marked in \textbf{bold} and \underline{underlined}, respectively.}
\label{tab:sts_results}
\vspace{-1.5ex}
\resizebox{\linewidth}{!}{ 
\begin{tabular}{ccccccccccccccccc}
\hline
 &
   &
   &
  \multicolumn{2}{c}{ATEC} &
  \multicolumn{2}{c}{BQ} &
  \multicolumn{2}{c}{LCQMC} &
  \multicolumn{2}{c}{PAWSX} &
  \multicolumn{2}{c}{STSB} &
  \multicolumn{2}{c}{AFQMC} &
  \multicolumn{2}{c}{QBQTC} \\ \hline
Metric       & \#Data & \#Param. & Pear.  & Spear. & Pear.  & Spear. & Pear.  & Spear. & Pear.  & Spear. & Pear.  & Spear. & Pear.  & Spear. & Pear.  & Spear. \\ \hline
BGE-ft(326M)      & 0.3M   & 326M     & \underline{0.3827} & \underline{0.4126} & 0.5388 & \textbf{0.5982} & 0.5683 & \underline{0.6531} & \underline{0.2982} & \textbf{0.3127} & 0.6648 & 0.6717 & \underline{0.3492} & \underline{0.3774} & 0.1982 & 0.2049 \\
RocketQAv2(326M)   & 0.3M   & 326M     & 0.1971 & 0.2362 & 0.3815 & 0.3962 & 0.5368 & 0.6089 & 0.1687 & 0.1558 & 0.5662 & 0.5894 & 0.1945 & 0.2381 & \underline{0.2325} & 0.2180 \\
SGPT(7B)         & 0.3M   & 0.4M     & 0.3045 & 0.3173 & 0.5135 & 0.5241 & 0.4715 & 0.4767 & 0.1842 & 0.1653 & 0.5973 & 0.5842 & 0.3033 & 0.3077 & 0.1717 & 0.1736 \\
Udever(7B)       & 0.3M   & 89M      & 0.3328 & 0.3602 & \underline{0.5389} & 0.5531 & 0.5369 & 0.5819 & 0.2041 & 0.2063 & 0.6509 & 0.6601 & 0.3177 & 0.3246 & 0.2088 & 0.2102 \\
LLaRRA(7B)       & 0.3M   & 89M      & 0.343  & 0.3575 & 0.5233 & 0.5369 & \underline{0.5698} & 0.5997 & 0.2113 & 0.2063 & \underline{0.6910} & \underline{0.7001} & 0.3046 & 0.3238 & 0.2127 & \underline{0.2254} \\
\textbf{D2LLM}(7B)        & 0.3M   & 89M      & 0.3731 & 0.3994 & 0.5487 & 0.5674 & 0.6210 & \textbf{0.6589} & \textbf{0.3038} & \underline{0.2883} & 0.7273 & 0.7194 & 0.3676 & 0.3858 & 0.2749 & 0.2850 \\ 
\textbf{D2LLM-ft}(7B)   & 0.3M    & 89M      & \textbf{0.4603} &  \textbf{0.4759}  &\textbf{0.5589} & \underline{0.5705}   &  \textbf{0.6233}	 &   0.6475  & 0.2145 &  0.2573   &   \textbf{0.7346} &  \textbf{0.7729}  & \textbf{0.3891} & \textbf{0.3966} & \textbf{0.2756}  &  \textbf{0.2933}  \\
Improv.(\%)  & N/A   & N/A      & 20.27\% &  15.34\%  &3.71\% & -4.63\%   &  9.39\%	 &   0.89\%  & 1.88\% &  -7.80\%  &   6.31\% &  10.40\%  & 11.43\% & 5.09\% & 18.54\%  &  30.12\%  \\
\hline
BGE(326M)          & 100M   & 326M     & 0.4716 & 0.4785 & 0.6001 & 0.6224 & 0.6924 & 0.7249 & 0.3001 & 0.3584 & 0.7765 & 0.7763 & 0.4100 & 0.4253 & 0.2203 & 0.2424 \\
LLM-be(7B)         & N/A    & N/A      & 0.2339 & 0.2178 & 0.3049 & 0.3007 & 0.4484 & 0.4507 & 0.1803 & 0.1676 & 0.5761 & 0.5767 & 0.1762 & 0.1837 & 0.1153 & 0.1147 \\
LLM-ce(7B)      & N/A    & N/A      & 0.3670 & 0.4152 & 0.4432 & 0.4770 & 0.6224 & 0.7164 & 0.3125 & 0.4365 & 0.7453 & 0.7680 & 0.3643 & 0.3986 & 0.3355 & 0.3491 \\ 
LLM-ce-ft(7B) & 0.3M    & 89M      & 0.4816 & 0.4898 & 0.5868 & 0.5991 & 0.6205 & 0.7147 & 0.1978 & 0.4002 & 0.7873 & 0.8172 & 0.4284 & 0.4254 & 0.3414 & 0.3516 \\ \hline
\end{tabular}}
\vspace{-3ex}
\end{table*}

\section{Experiments}
In this section, we evaluate D2LLM's effectiveness on three tasks: Natural Language Inference (\textbf{NLI}), Semantic Textual Similarity (\textbf{STS}), and Information Retrieval (\textbf{IR}), with a focus on Chinese. The former two are symmetric search tasks, and the last is asymmetric. Details on datasets and evaluation metrics are provided in Appendix~\ref{app:dataset}. For NLI and STS, we use 0.3 million training samples, and 0.9 million for IR. The proposed method (denoted as D2LLM\footnote{The implementation deltals is presented in Appendix~\ref{app:implementaion}}) is compared with five state-of-the-art baselines: \textbf{BGE}~\cite{xiao2023c}, \textbf{RocketQAv2}~\cite{ren2021pair}, \textbf{SGPT}~\cite{muennighoff2022sgpt}, \textbf{Udever}~\cite{zhang2023language}, and \textbf{LLaRA}~\cite{li2023making}. BGE and RocketQAv2 are BERT-style bi-encoders; BGE uses pretraining and finetuning, while RocketQAv2 distills from a cross-encoder, updating both the student and teacher. The remaining three finetune GPT-style LLMs into bi-encoders. For more details, please refer to Appendix~\ref{app:baselines}. The total number of parameters for each method is shown in parentheses after it. To ensure a fair comparison, all baseline methods have a number of tunable parameters (denoted as \#Param. in Tables~\ref{tab:nli_results}-\ref{tab:sts_results}) equal to, or greater than, those in D2LLM, with the exception of SGPT, which regards the use of BitFit finetuning~\cite{zaken2022bitfit} as its notable advantage. All methods, except SGPT, incorporate the hard negatives from the previous section, as SGPT's original training does not involve such negatives. It's worth noting that BGE relies on extensive data for pretraining and finetuning, while other methods only use the small dataset mentioned above for finetuning. As the pretrained BGE model was not accessible, we opted for Chinese-roberta-large-326M~\cite{cui2021pre} and finetuned it using BGE's method with our dataset, referring to this version as \textbf{BGE-ft}. The original BGE is still retained as a reference. We also use Chinese-roberta-large as the base for RocketQAv2. For the other methods, we select Qwen-7B-Chat~\cite{bai2023qwen} as the base model. Additionally, the performance of the cross-encoder teacher model based on Qwen-7B-Chat (i.e., \textbf{LLM-ce}) is presented, as well as a bi-encoder (i.e., \textbf{LLM-be}) that generates sentence embeddings by mean-pooling token embeddings from the last layer of Qwen-7B-Chat, in order to gauge D2LLM's improvement over the untrained LLM-be and its proximity to the teacher LLM-ce's performance.
\begin{figure*}[t]
\begin{minipage}[c]{0.34\linewidth}
    \centering
    \includegraphics[width=1.0
    \textwidth]{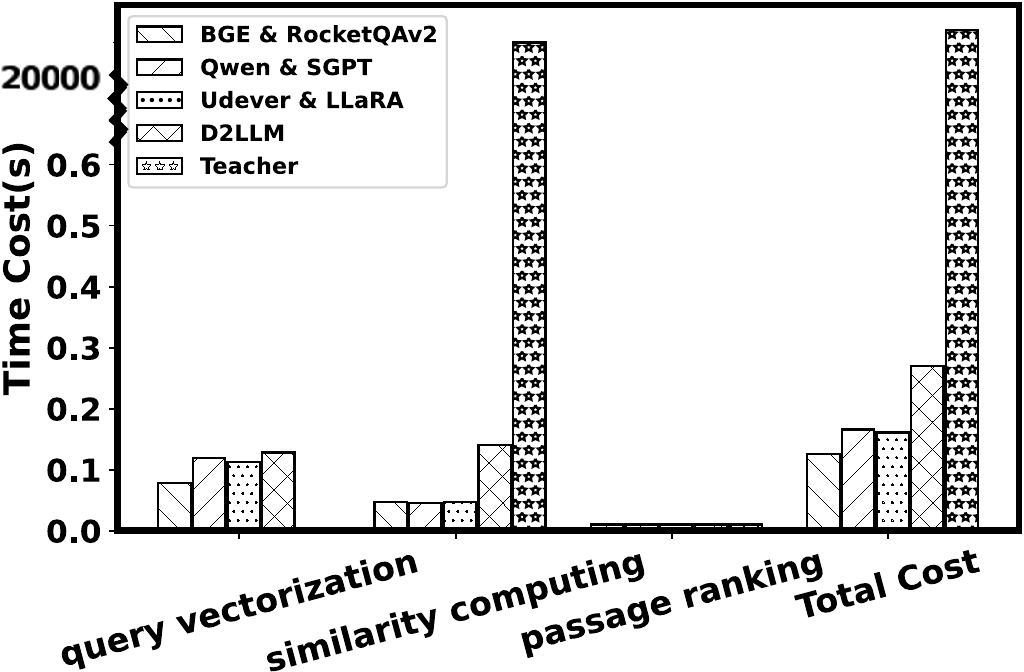}
    \vspace{-4ex}
    \caption{Runtime Analysis.}
    \label{fig:runtime}
\end{minipage}
\hfill
\begin{minipage}[c]{0.64\linewidth}
\centering
\captionof{table}{Ablation Study on the NLI Task.}
\vspace{-1.5ex}
\label{tab:ablation_study}
\resizebox{\linewidth}{!}{ 
\begin{tabular}{ccccccccc|c}
\hline
              & \multicolumn{4}{c}{OCNLI}         & \multicolumn{4}{c|}{CMNLI} &\multirow{2}{*}{\begin{tabular}[c|]{@{}c@{}}Average  \\ difference\end{tabular}}        \\ \cline{1-9}
Methods       & ACC    & AP     & Prec.  & Recall & ACC    & AP     & Prec.  & Recall \\ \hline
$-$CI$+$CL & 0.7572 & 0.7791 & 0.7411 & 0.7887 & 0.7755 & 0.8471 & 0.7692 & 0.8018 & -3.59\%  \\
$-$RI\_PH   & 0.7293 & 0.7658 & 0.7106 & 0.7589 & 0.7567 & 0.8129 & 0.7433 & 0.7885 & -6.57\% \\
$-$RI\_HI   & 0.7375 & 0.7726 & 0.7390 & 0.7711 & 0.7721 & 0.8194 & 0.7609 & 0.7990 & -4.92\%\\
$-$FI      & 0.7666 & 0.8012 & 0.7518 & 0.8006 & 0.7939 & 0.8542 & 0.7771 & 0.8056 & -2.17\% \\
$-$PMA$+$mean    & 0.7734 & 0.7997 & 0.7586 & 0.8022 & 0.7954 & 0.8611 & 0.7823 & 0.8087 & -1.71\%\\
$-$PMA$+$[EOS]    & 0.7739 & 0.8023 & 0.7604 & 0.8025 & 0.7958 & 0.8642 & 0.7845 & 0.8107 &-1.51\% \\
$-$IEM$+$cos     & 0.7461 & 0.7886 & 0.7224 & 0.8025 & 0.7867 & 0.8377 & 0.7682 & 0.7921 & -3.83\%\\ 
D2LLM-1.8B & 0.6907 & 0.7399 & 0.6769 & 0.6261 & 0.7102 & 0.7947 & 0.7107 & 0.5840 & -14.76\%\\\hline
\textbf{D2LLM}         & 0.7889 & 0.8145 & 0.7736 & 0.8149 & 0.8014 & 0.8759 & 0.7960 & 0.8241 & -\\ \hline
\end{tabular}}
\end{minipage}
\vspace{-2ex}
\end{figure*}
\subsection{Results on NLI}
We first investigate the performance of all methods for NLI. Table~\ref{tab:nli_results} shows that D2LLM outshines all competitors trained on the 0.3M sample set across all metrics and all testing datasets. Notably, it surpasses LLaRA, the second-best method, by a significant margin of 14.39\%\footnote{We compute the relative gain in this paper, which is defined as the ratio of the difference between a new value and a reference value to the reference value itself.} on average and exceeds BGE, which was finetuned on 100M relevant samples, by 6.45\%. Furthermore, D2LLM effectively narrows the gap between the intact bi-encoder LLMs (LLM-be) and cross-encoder LLMs (LLM-ce). While the original LLM-be falls short as a bi-encoder due to the mismatch between text generation and embedding, the cross-encoder-based teacher model LLM-ce excels by leveraging LLMs' ability to synthesize information from sentence pairs. Our distillation approach successfully transfers knowledge from the teacher to the student, transforming the initially ineffective LLM-be into a proficient NLI instrument. Additionally, SGPT, Udever, and LLaRA outperform the base LLM-be, highlighting that contrastive finetuning boosts LLM-be's capacity, albeit not as effectively as D2LLM's distillation. Among these, LLaRA and Udever benefit more from hard negatives in contrastive finetuning than in-batch negatives based SGPT, with LLaRA's token-level contrastive learning proving more advantageous than Udever's sentence-level approach. Furthermore, a comparison between BGE and BGE-ft reveals that BGE's impressive performance is likely attributed to its extensive pretraining and finetuning datasets. Finally, the performance of RocketQAv2 falls short of expectations, likely due to having fewer hard negatives per sample in our experiments (i.e., 8) compared to the original setting (i.e., 32 or 127); the efficacy of listwise distillation in RocketQAv2 depends on the hard negatives. Despite its outstanding performance, D2LLM still exhibits poor performance on certain samples, please refer to \ref{app:bad_case} for details.



\subsection{Results on STS and IR}
Moving on to the STS task, Table~\ref{tab:sts_results} shows that D2LLM outperforms other methods trained on the same dataset in most cases (10 out of 14). However, BGE-ft occasionally exceeds D2LLM, and the original BGE maintains a consistent lead. Notably, even the teacher model, LLM-ce, falls behind BGE, shedding light on D2LLM's less-than-optimal results. However, It's important to point out that the teacher model LLM-ce was not specifically finetuned for STS. To address this, we finetune the teacher model with the same training set for the STS domain using LoRA (see Appendix~\ref{app:teacher_tune} for more details), resulting in the variant \textbf{LLM-ce-ft}. The finetuning on merely 0.3M data yields an average improvement of 7.17\% for the teacher, an effect showed in the last two rows in Table~\ref{tab:sts_results}. Building upon LLM-ce-ft, we train the student model, denoted as \textbf{D2LLM-ft}, which shows an increase of 1.69\% over the original D2LLM. 
Furthermore, now D2LLM-ft significantly outshines other methods trained on the same 0.3M sample set, by a margin of at least 17.42\% on average. This confirms that, despite the initial underperformance on a task, the strong adaptability of LLMs means that finetuning with a relatively small dataset can substantially enhance both the teacher's and subsequently the student's performance. On the other hand, while BGE shows comparable or marginally better results, it does so at the expense of finetuning on a massive 100M data corpus specific to semantic search tasks, suggesting that adopting a smaller model like BGE (326M parameters) demands a large quantity of pertinent data. Regarding the IR task, its results are similar to those of the prior two tasks, and so we deferred their detailed discussion to Appendix~\ref{app:ir_results}. Similar to the NLI task, D2LLM struggles with some specific IR samples, as detailed in \ref{app:bad_case}.


\subsection{Runtime Analysis}
In this section, we evaluate the runtime of all methods, which can be further divided into the time for query vectorization, relevance scoring, and passage ranking. The results of all methods for the dataset T2Retriever are directed in Figure~\ref{fig:runtime}. For bi-encoder-based methods, since passage embedding vectors are pre-computed and stored in a database, we exclude that preprocessing time and instead measure only the runtime for query vectorization and cosine similarity calculation. The proposed D2LLM follows a similar vectorization step, but computes similarity using a multi-layer perceptron (MLP). For cross-encoder-based method LLM-ce, each passage must be concatenated with the query and processed individually through the model, resulting in a significantly higher runtime for relevance scoring compared to bi-encoders. D2LLM demonstrates only a marginally increased relevance-scoring time relative to bi-encoders, attributable to the more complex computations within the MLP compared to cosine similarity. In summary, D2LLM markedly improves upon the cross-encoder-based teacher model's efficiency while enhancing the accuracy of the bi-encoder benchmark, effectively balancing efficiency and accuracy.

\subsection{Ablation Studies}
Due to the space limitation, we only briefly overview our ablation studies here. For a detailed account, please see Appendix~\ref{app:impact_module}. We first explore the individual contributions of different losses and modules to the performance of D2LLM. Insights gleaned from Table~\ref{tab:ablation_study} reveal that: 1) The integration of contrastive, rank, and feature imitation processes is critical to D2LLM's success; these processes effectively distill diverse facets of knowledge from the teacher model. 2) PMA proves indispensable and cannot be aptly substituted with mean pooling or the use of the [EOS] token within the LLM; these alternatives are either overly simplistic or misaligned with the inherent design of the LLM. 3) the IEM is necessary for capturing the intricate dynamics of query-passage relationships, which cosine similarity alone fails to encapsulate adequately. 4) The size of the teacher LLM positively influences the performance of the student D2LLM, with larger teachers leading to more capable students. Furthermore, we find in Appendix~\ref{app:branch} that the IEM with dual branches can handle both symmetric and asymmetric semantic tasks, outperforming single-branch variants that struggle outside their specialized domains. Finally, we perform a sensitivity analysis of the weights in the overall loss~\eqref{eq:loss_overall} in Appendix~\ref{app:weight}, and use the selected values in all our experiments.

\section{Conclusion}
This paper presents D2LLM, an innovative model distilling knowledge from an LLM teacher to construct an efficient student for semantic search. With a deep and nuanced understanding of its teacher, D2LLM utilizes specially designed modules and losses to encapsulate the teacher's prowess in a more compact form. Our experimental findings reveal that D2LLM successfully synthesizes the high accuracy associated with cross-encoders and the operational efficiency of bi-encoders. 

\section{Ethical Considerations}

Our research is of a fundamental nature and is not anticipated to have significant social implications. We have utilized exclusively open-source datasets, which ensures transparency and adherence to ethical standards in data usage. Moreover, the accessibility of these datasets facilitates replicability and scrutiny by the broader research community, aligning with ethical research practices. However, we acknowledge the responsibility that accompanies the development of any semantic search technology, given its potential influence on information access and dissemination. We encourage ongoing dialogue and ethical considerations in the application of our research findings.

\section{Limitations}
While D2LLM presents a significant advancement in semantic search, it is not without limitations. Firstly, the IEM offers greater flexibility than cosine similarity in capturing semantic nuances, but it may still omit some intricate details grasped by the original teacher LLM. While enhancing the IEM's complexity could improve performance, this can come at the expense of efficiency. Secondly, the model incorporates three weight parameters in the combined loss function~\eqref{eq:loss_overall}, and optimally tuning these hyperparameters from the data remains a complex yet potentially rewarding challenge. Finally, training D2LLM demands substantial computational resources due to its reliance on large language models, which could be prohibitive for those with limited computational means.

\section{Acknowledgements}
We would like to thank Ant Group for their support for this work.
This work was supported in part by National Natural Science Foundation of China (No. 62072182 and No. 92270119).

\bibliography{anthology,custom, ./bibs/general.bib}
\appendix

\section{Datasets}
\label{app:dataset}
We consider three tasks to verify the usefulness of all methods, including Natural Language Inference (NLI), Semantic Textual Similarity (STS), and Information Retrieval (IR). The training and testing data for each task are listed below. Note that all testing data are taken from the comprehensive benchmark CMTEB (Chinese Massive Text Embedding Benchmark).  
\begin{itemize}[leftmargin=*, topsep=1pt,itemsep=1pt,partopsep=1pt,parsep=1pt]
    \item \textbf{NLI}:~For the NLI task, models are tasked to discern the presence of an entailment relationship between pairs of sentences. The training data for this task involves the 0.3M data from SNLI-zh\footnote{https://huggingface.co/datasets/shibing624/snli-zh}, NLI-zh\footnote{https://huggingface.co/datasets/shibing624/nli\_zh}, specifically including the datasets named ATEC, BQ, LCQMC, and PAWSX. The testing datasets are OCNLI and CMNLI. Performance metrics include Accuracy, Average Precision (AP), Precision, and Recall.
    \item \textbf{STS}:~For the STS task, the objective is to predict the degree of similarity between sentence pairs, with a higher predicted score indicating greater similarity. The training data is the same as the above NLI task. The testing data involves ATEC, BQ, LCQMC, PAWSX, STSB, AFQMC, and QBQTC. Pearson and Spearman correlation coefficients serve as evaluation metrics.
    \item \textbf{IR}: For the IR task, each dataset comprised a corpus, a set of queries, and an associated mapping of each query to the relevant documents within the corpus. The aim was to accurately identify these pertinent documents for each query. The training data are randomly sampled from T2Ranking\footnote{https://github.com/THUIR/T2Ranking}, DuReader\footnote{https://github.com/baidu/DuReader}, cMedQA2\footnote{https://github.com/zhangsheng93/cMedQA2}, and mMARCO\footnote{https://huggingface.co/datasets/unicamp-dl/mmarco}. Concretely, we sample 50\% from T2Ranking, 80\% from DuReader, 80\% from cMedQA2, and 35\% from mMARCO, and finally compose a training dataset of 0.9M. The testing data involves T2Retrieval, DuRetrieval, CovidRetrieval, CmedqaRetrieval, MedicalRetrieval, and MMarcoRetrieval. To evaluate the retrieval effectiveness, each method retrieves and ranks the top-10 passages for each query, and MRR@10 and Recall@10 are then utilized as the metrics.
\end{itemize}

\section{Implementation Details}
\label{app:implementaion}

Our D2LLM model is built upon PyTorch and DeepSpeed, using Qwen-7B-Chat as the teacher and the base LLM for the student due to its effectiveness with Chinese data. The model uses a batch size of 32, with each query having 8 hard negatives assigned. The PMA module features 32 heads, and the IEM includes two single-layer MLPs with ReLU activations—the first with an input size of 8192 and output of 512, and the second with consistent 512 dimensions for both. We set $\alpha = 1$, $\beta = 0.3$, and $\gamma = 0.1$ in all our experiments, unless otherwise specified, based on the observations in Appendix~\ref{app:weight}. The AdamW optimizer is used with a learning rate of 1e-4, including a warm-up over 0.2 epochs, and training is halted early upon model convergence. LoRA adjustments are made with a rank of 8, while mixed-precision training and gradient checkpointing minimize memory usage. Training runs on 8 NVIDIA A100 GPUs with 80GB each. 

For the consumption of computing resources, note that the number of tunable parameters in D2LLM, which amounts to 89 million including the modifications for PMA and IEM, remains on par with Udever and LLaRA. The consistent parameter count across these methods is due to our choice of using the LoRA adaptation technique for all models, and the foundational model is the same for D2LLM, Udever, and LLaRA, leading to a comparable computational cost for these LLM-based approaches.
Besides, to optimize GPU memory utilization, we engage in a two-stage approach: initially, we utilize the Teacher LLM to infer and store logits, relevant for contrastive and rank imitation losses, as well as similarity matrices necessary for feature imitation. These logits and similarity matrices, with relatively small memory footprints, correlate only with selected positives and hard negatives for each sample. Next, during the subsequent training phase, the student model can learn from these pre-saved components without the simultaneous GPU presence of the teacher model. In practice, D2LLM training culminates in approximately 10 and 22 hours of training time for the symmetric (i.e., NLI and STS) and asymmetric search tasks (i.e., IR) respectively. By contrast, the training durations for LLM-based bi-encoders, such as LLaRA, hover around 9 and 20 hours for the same tasks. This indicates that the resource usage is nearly equivalent between these methods, with D2LLM introducing only a minimal additional computational burden. To implement D2LLM with diminished resource utilization, a reduction in the number of tunable parameters could be beneficial—potentially by lowering the rank value in LoRA or by adopting other parameter-efficient fine-tuning methods like QLoRA~\cite{dettmers2023qlora}, and investigating these resource-conserving alternatives will be a focus of our future work.

\begin{table*}[t]
\centering
\caption{Results for IR, with the best-performing method and the second-best results marked in \textbf{bold} and \underline{underlined}, respectively.}
\label{tab:ir_results}
\resizebox{0.95\linewidth}{!}{ 
\begin{tabular}{ccccccccccccccc}
\hline
 &  &  & \multicolumn{2}{c}{T2Retrieval} & \multicolumn{2}{c}{DuRetrieval} & \multicolumn{2}{c}{mMARCORetrieval} & \multicolumn{2}{c}{cMedQARetrieval} & \multicolumn{2}{c}{CovidRetrieval} & \multicolumn{2}{c}{MedicalRetrieval} \\ \hline
Metric & \#Data & \#Param. & MRR & Recall & MRR & Recall & MRR & Recall & MRR & Recall & MRR & Recall & MRR & Recall \\ \hline
BGE-ft(326M) & 0.9M & 326M & \underline{0.8739} & \underline{0.7659} & \underline{0.9125} & \underline{0.8546} & \textbf{0.6908} & \textbf{0.8363} & 0.2208 & 0.2638 & \textbf{0.5828} & \textbf{0.7363} & 0.4561 & 0.5490 \\
RocketQAv2(326M) & 0.9M & 326M & 0.6670 & 0.5579 & 0.6502 & 0.5079 & 0.5012 & 0.6920 & 0.1995 & 0.2321 & 0.4032 & 0.5822 & 0.2979 & 0.3851 \\
SGPT(7B) & 0.9M & 0.4M & 0.7408 & 0.6026 & 0.7762 & 0.6805 & 0.5516 & 0.7307 & 0.2454 & 0.3249 & 0.4411 & 0.6109 & 0.3266 & 0.4121 \\
Udever(7B) & 0.9M & 89M & 0.8653 & 0.7358 & 0.8905 & 0.8213 & 0.6327 & 0.7898 & 0.3145 & \underline{0.3903} & 0.5124 & 0.6742 & 0.4346 & 0.5210 \\
LLaRA(7B) & 0.9M & 89M & 0.8412 & 0.7362 & 0.8777 & 0.8083 & 0.6511 & 0.8003 & \underline{0.3221} & 0.3886 & 0.5250 & 0.6793 & \underline{0.4612} & \underline{0.5632} \\
\textbf{D2LLM}(7B) & 0.9M & 89M & \textbf{0.8893} & \textbf{0.7719} & \textbf{0.9162} & \textbf{0.8608} & \underline{0.6723} & \underline{0.8221} & \textbf{0.3501} & \textbf{0.4028} & \underline{0.5639} & \underline{0.7121} & \textbf{0.4991} & \textbf{0.6021} \\ 
Improv.(\%) & N/A & N/A & 1.76\% & 0.78\% & 0.41\% & 0.73\% & -2.68\% & -1.70\% & 8.69\% & 3.20\% & -3.24\% & -3.29\% & 8.22\% & 6.91\% \\ \hline
BGE(326M) & 100M & 326M & 0.9094 & 0.8084 & 0.9345 & 0.8851 & 0.7583 & 0.8934 & 0.4349 & 0.4452 & 0.6587 & 0.8246 & 0.5504 & 0.6660 \\
LLM-be(7B) & N/A & N/A & 0.1843 & 0.1098 & 0.3331 & 0.2178 & 0.0541 & 0.1011 & 0.0456 & 0.0547 & 0.2630 & 0.4131 & 0.0462 & 0.076 \\
LLM-ce(7B) & N/A & N/A & - & - & - & - & - & - & - & - & - & - & - & - \\ \hline
\end{tabular}}
\end{table*}

\section{Baselines}
\label{app:baselines}

The five benchmark methods are summarized below:
\begin{itemize} [leftmargin=*, topsep=1pt,itemsep=1pt,partopsep=1pt,parsep=1pt]
    \item \textbf{BGE}~\cite{xiao2023c} is a BERT-style bi-encoder. It involves three training stages. First, the model pretrained with massive data using MAE. Next, it is finetuned with unlabeled and labeled data separately. 
    \item \textbf{RocketQAv2}~\cite{ren2021rocketqav2} is also BERT-style bi-encoder. It is distilled from a BERT-style cross-encoder via dynamic listwise distillation. This technique enables joint update of both the teacher and student. Particularly in our experiments, we initialize both the student and teacher as a pretrained Chinese-roberta-large model\footnote{https://huggingface.co/hfl/chinese-roberta-wwm-ext-large}. 
    \item \textbf{SGPT}~\cite{muennighoff2022sgpt} exploits both bi and cross-encoder architectures to enhance GPT-style LLMs for semantic search. Here, we only utilize the bi-encoder variant due to its efficiency. It incorporates BitFit finetuning of the LLM and position-weighted mean pooling to generate an overall embedding for a query or a passage.
    \item \textbf{Udever}~\cite{zhang2023language} also aims to modify GPT-style LLMs for text embedding. Udever introduces a special token appended to the end of the sentences and trains this token to summarize the sentences via sentence-level contrastive learning.
    \item \textbf{LLaRA}~\cite{li2023making} is similar to Udever, but employs token-level contrastive learning to further refine sentence embeddings. In particular, for Udever and LLaRA, we set the rank of the low-rank adapters (LoRA) to 40, in order to guarantee the number of trainable parameters in these two methods is the same as that in D2LLM.
\end{itemize}

\section{Teacher Model Finetuning}
\label{app:teacher_tune}
To finetune the teacher model for a specific task, we incorporate all positive sentence pairs and select an equal number of hard negatives (cf. Section~\ref{ssec:ci}), maintaining a balanced ratio of 1:1 between positives and negatives. Depending on the task at hand, we utilize prompts suited for either symmetric or asymmetric searches to structure the finetuning dataset, composing inputs of prompted sentence pairs and outputs of binary responses "yes" or "no".

We typically opt for LoRA finetuning, where we set the rank within LoRA to 32. This specific rank setting is chosen to align the number of learnable parameters with those used in other comparable methods.

\section{Results on IR}
\label{app:ir_results}
In this section, we check the performance of all methods for the IR task. As shown in Table~\ref{tab:ir_results}, D2LLM again outperforms other methods trained on the same dataset for the majority of the time (8 out of 12 instances), despite not having a teacher model finetuned specifically for this task. Although we lack performance data for the teacher model LLM-ce due to its cross-encoder design being impractically slow for real-world retrieval tasks, D2LLM proves to be an effective surrogate. It strikes a balance between accuracy and efficiency, making it suitable for practical applications. Nonetheless, it's worth noting that BGE achieves the highest performance, a likely result of its extensive training on data pertinent to IR.

\section{Ablation Studies}
\label{app:ablation}

In this section, we conduct a series of ablation studies to further show the competency of the proposed D2LLM.

\subsection{Impact of Losses and Modules}
\label{app:impact_module}

Our analysis focuses on the efficacy of various losses and modules detailed in Section~\ref{sec:method}. Below are the specific modifications we tested, with the corresponding results shown in Table~\ref{tab:ablation_study}:
\begin{itemize} [leftmargin=*, topsep=1pt,itemsep=1pt,partopsep=1pt,parsep=1pt]
    \item\textbf{$-$CI$+$CL}: We replaced the proposed CI loss in Section~\ref{ssec:ci} with the standard contrastive loss. The CI loss amounts to the standard contrastive loss by setting the score $s_{ij}^\gT$ given by the teacher to 1. As expected, the resulting performance deteriorates by 3.59\%, since the standard contrastive loss is sensitive to positives concealed within the hard negative set $\sD^-_H$.
    \item\textbf{$-$RI\_PH}: Omitting the rank imitation loss for positives and hard negatives (Equation~\eqref{eq:loss_ri_ph}) led to a 6.57\% reduction in performance. This underscores the value of the student model mirroring the teacher in ranking these critical pairs.
    \item\textbf{$-$RI\_HI}: The removal of the rank imitation loss for hard and easy negatives (Equation~\eqref{eq:loss_ri_hi}) resulted in a 4.92\% performance drop. This supports our initial argument (Section~\ref{ssec:ri}) that distinguishing these sample sets is key for robust student model training.
    \item\textbf{$-$FI}: Excluding the feature imitation loss incurred a 2.17\% loss in performance, highlighting the role of feature distillation in transferring a broader spectrum of knowledge from the teacher to the student model.
    \item\textbf{$-$PMA$+$mean}: Replacing Pooling by Multihead Attention (PMA) with mean pooling led to a 1.71\% decrease in performance. This result emphasizes the superior flexibility of the learnable PMA compared to the static mean pooling. 
    \item\textbf{$-$PMA$+$[EOS]}: Forgoing PMA in favor of using the [EOS] token as a sentence-wide embedding, and applying contrastive finetuning, caused a 1.51\% performance downturn. The shift in training objective strays from the original purpose of the [EOS] token in the pretrained LLM, thus not fully capitalizing on the LLM's capabilities.
    \item\textbf{$-$IEM$+$cos}: Substituting the Interaction Emulation Module (IEM) with cosine similarity, akin to original bi-encoders, led to a 3.83\% decline in performance. This change buttresses our assertion that the MLP is integral to modeling complex sentence relationships more effectively than cosine similarity alone.
    \item\textbf{D2LLM-1.8B}: Scaling down the teacher LLM from 7B to 1.8B exhibited a 14.76\% decrease in performance. This suggests that the capacity of the teacher LLM is a key determinant in the effectiveness of the student D2LLM, with the findings indicating that larger teacher models engender more proficient student models.
\end{itemize}

\begin{table}[t]
\caption{Performance on individual or mixed data type via IEM and cosine similarity.}
\label{tab:mixed_training}
\resizebox{\linewidth}{!}{ 
\begin{tabular}{ccccccc}
\hline
                 & \multicolumn{2}{c}{OCNLI} & \multicolumn{2}{c}{CMNLI} & \multicolumn{2}{c}{T2Retrieval} \\ \hline
Metric           & ACC         & AP          & ACC         & AP          & MRR            & Recall         \\ \hline
D2LLM-cos-sym   & 0.7461      & 0.7886      & 0.7867      & 0.8377      & 0.2791              & 0.2021              \\
D2LLM-cos-asym & 0.4913      & 0.4704     & 0.5003           & 0.5082           & 0.8072         & 0.7037         \\
D2LLM-cos-mixed   & 0.7035      & 0.7524      & 0.7593      & 0.8024      & 0.7771              & 0.6802              \\ \hline
D2LLM-sym   & 0.7905      & 0.8226      & 0.8084      & 0.8839      & 0.2622              & 0.1650              \\
D2LLM-asym & 0.5138      & 0.4893     & 0.5144           & 0.5003           & 0.8346         & 0.7218         \\
\textbf{D2LLM-dual}     & 0.7834      & 0.8017      & 0.7825      & 0.8778      & 0.8321         & 0.7059         \\ \hline
\end{tabular}}
\end{table}

\begin{figure*}[t]
    \centering 
        \includegraphics[width = 0.32\linewidth]{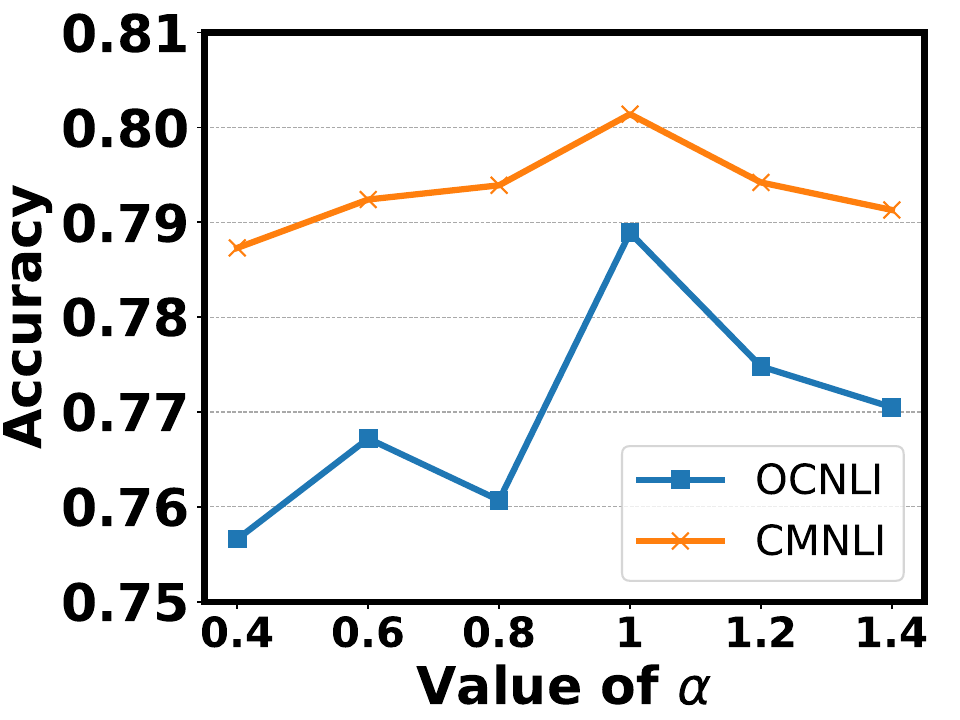}
        \includegraphics[width = 0.32\linewidth]{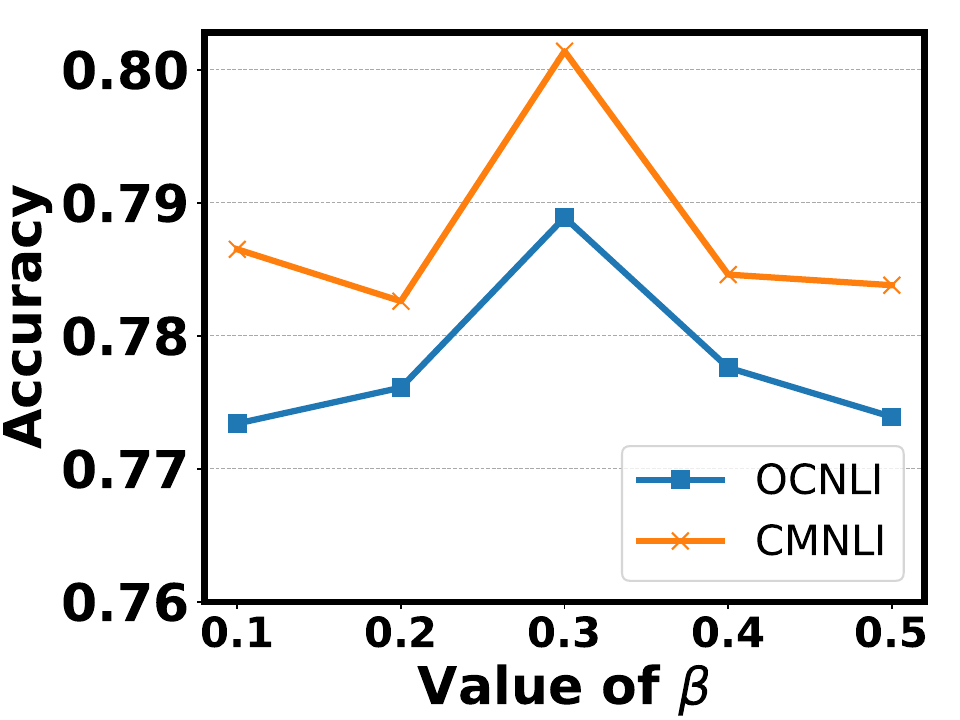}
        \includegraphics[width = 0.32\linewidth]{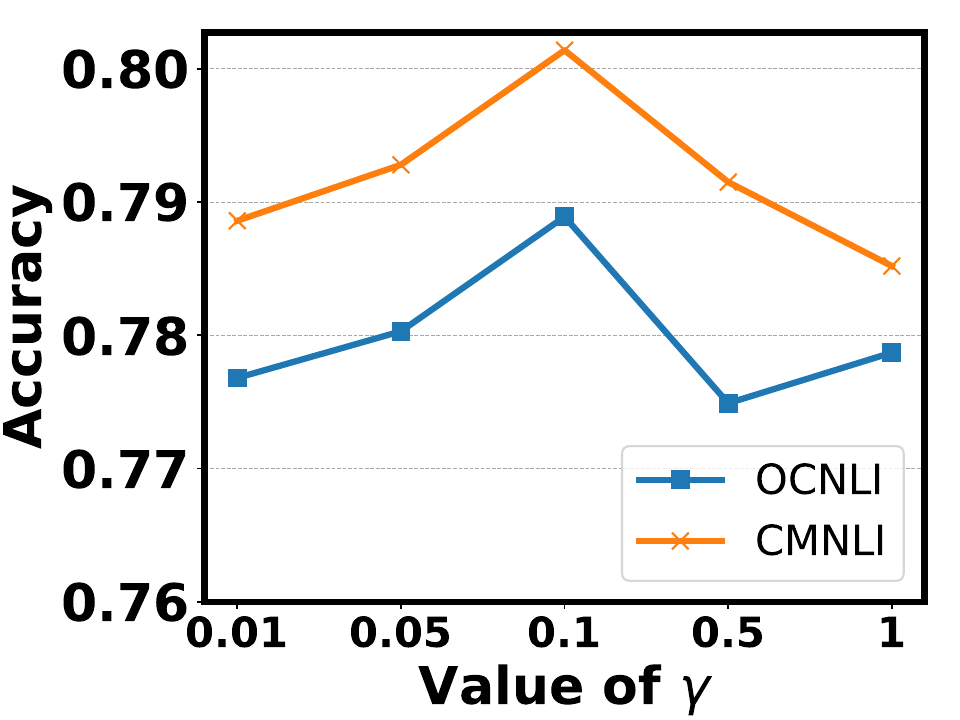}
    \caption{Effect of the hyperparameter $\alpha$, $\beta$ and $\gamma$ on OCNLI and CMNLI.}
    \label{param_sensi}
\end{figure*}
\begin{figure*}[t]
    \centering 
        \includegraphics[width = 0.32\linewidth]{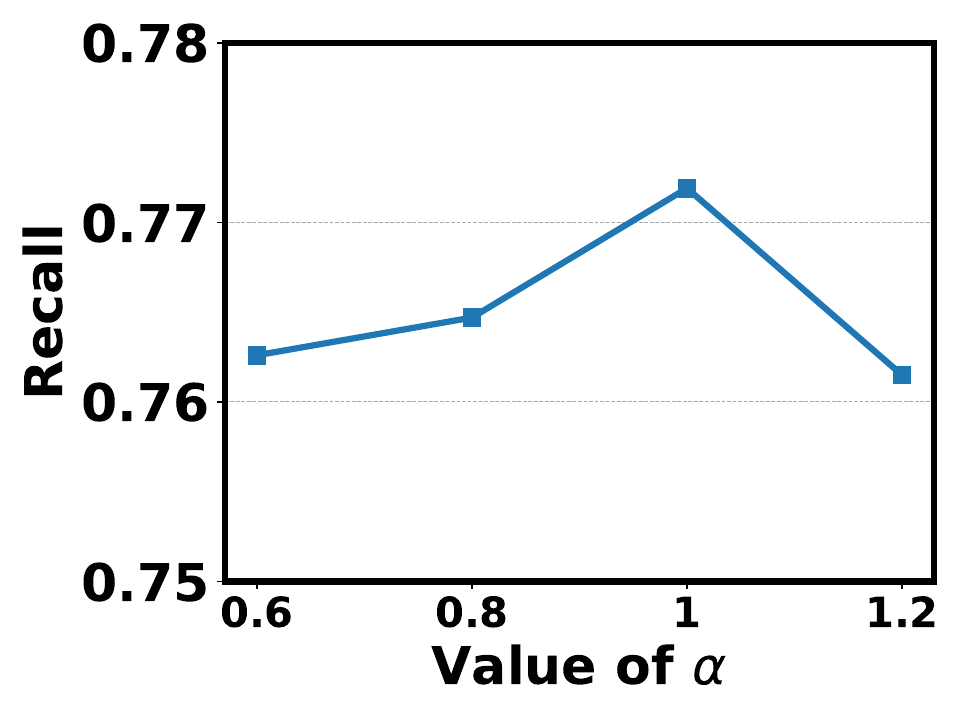}
        \includegraphics[width = 0.32\linewidth]{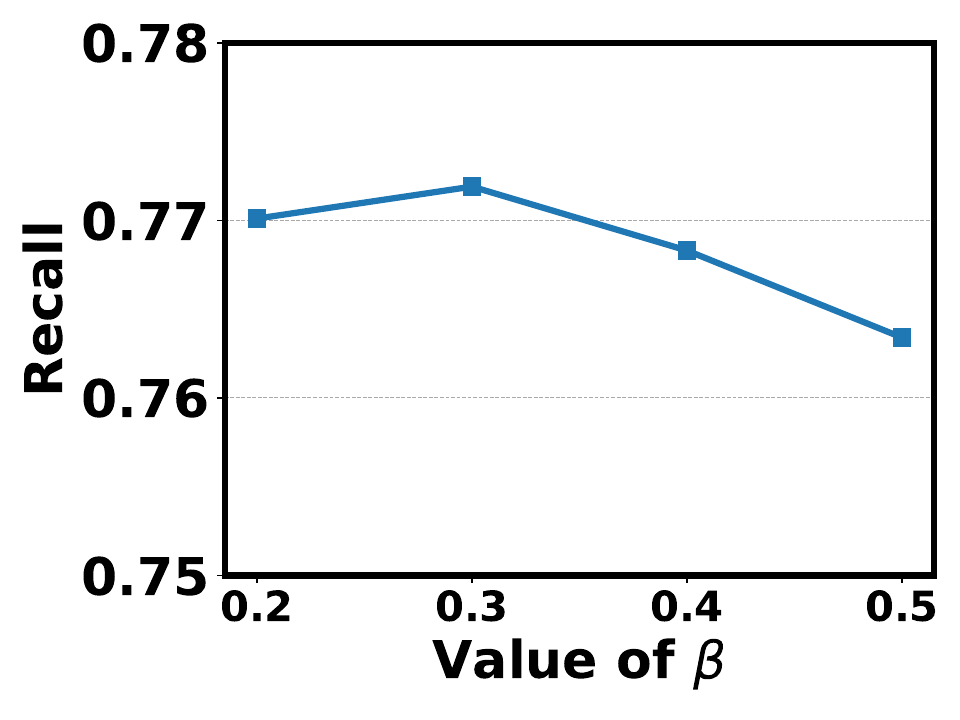}
        \includegraphics[width = 0.32\linewidth]{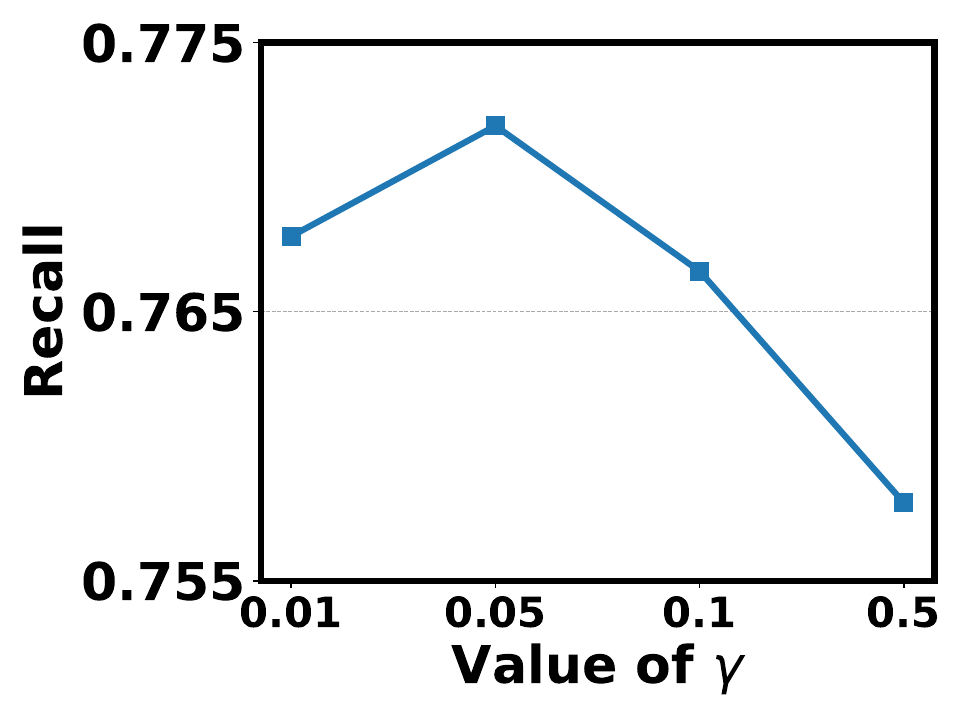}
    \caption{Effect of the hyperparameter $\alpha$, $\beta$ and $\gamma$ on T2Retrieval.}
    \label{param_sensi_t2}
\end{figure*}

\subsection{Impact of Single and Dual Branches in the IEM}
\label{app:branch}

We further investigated whether it is essential and beneficial to include two branches in the IEM. For this purpose, we selected two representative datasets from those listed in Appendix~\ref{app:dataset}: SNLI-zh for symmetric semantic relationships and T2Ranking for asymmetric semantic relationships. These datasets were employed to train a dual-branch model, designated as D2LLM-dual. For comparative analysis, we trained two separate models—D2LLM-sym and D2LLM-asym—each specializing in one of the two dataset types. We assessed the models' capabilities on test datasets that were aligned with either symmetric relations (OCNLI and CMNLI) or asymmetric relations (T2Retrieval). The results are summarized in Table~\ref{tab:mixed_training}.

It can be observed that D2LLM-dual achieves performance on par with D2LLM-sym for symmetric relationship tasks and with D2LLM-asym for asymmetric tasks. In stark contrast, D2LLM-sym experiences a notable performance drop in asymmetric contexts, while D2LLM-asym struggles similarly with symmetric tasks. These findings highlight the efficacy of the dual-branch IEM in D2LLM-dual, which is adept at tackling both symmetric and asymmetric semantic challenges. Thus, D2LLM-dual emerges as a versatile and unified model, proficiently capturing the nuanced interactions between sentences regardless of the semantic relationship type.

\subsection{Weight Sensitivity Analysis}
\label{app:weight}
We also investigate the impact of three hyperparameters $\alpha$, $\beta$, and $\gamma$, which represent the balance weights of $\gL^\text{CI}$, $\gL^\text{RI}_{PH}$, and $\gL^\text{RI}_{HI}$ respectively. In practice, to determine ($\alpha$, $\beta$, $\gamma$), a grid search is obviously impractical. Instead, we opt for a sequential method to choose the weight parameters. Specifically, we first obtain a locally optimal value for $\alpha$ focusing solely on the losses $\mathcal L^{CI}$ and $\mathcal L^\text{RI}_{PH}$. Next, by slightly expanding the tentative range of $\alpha$, we further incorporate the loss component $\mathcal L^\text{RI}_{HI}$ and adjust to identify an optimal pair ($\alpha$, $\beta$). We continue in a similar fashion to identify the optimal ($\alpha$, $\beta$, $\gamma$) within the vicinity of the previously determined ($\alpha$, $\beta$). Although this approach may not guarantee globally optimal parameters, it substantially conserves resources and has proven to yield satisfactory performance. We conduct experiments based on NLI and IR tasks and report the performance for OCNLI, CMNLI, and T2Retrieval.

We first depict ACC and Recall as functions of $\alpha$ in Figure~\ref{param_sensi} and Figure~\ref{param_sensi_t2}. It can be observed that a low $\alpha$ value correlates with suboptimal performance, indicating that the student model is inadequately leveraging the teacher's ranking capacity. As $\alpha$ is incremented to 1, we observe a progressive enhancement in performance. However, surpassing this threshold results in a decline, which suggests an optimal value at $\alpha=1$.

Regarding $\beta$, this parameter determines the attention on the loss term $\gL^\text{RI}_{HI}$. The empirical results, depicted in Figure~\ref{param_sensi} and Figure~\ref{param_sensi_t2}, reveal that a $\beta$ value of 0.3 yields the most significant augmentation in model performance.

Finally, we change the value of $\gamma$ to discern the importance of feature imitation. Figure~\ref{param_sensi} and Figure~\ref{param_sensi_t2} indicate that higher $\gamma$ leads to enhanced model performance up to a certain point. Nevertheless, an excessive prominence on it disrupts the equilibrium among the other loss functions, thereby impairing overall performance. 

Based on the above analysis, we note that the optimal hyperparameter combination remains similar across different tasks, suggesting that D2LLM generally does not require laborious task-specific hyperparameter adjustments, thereby aiding in resource conservation. Experimental results for both the NLI and IR tasks reveal that D2LLM's performance is relatively stable across a certain parameter range for the weight parameters ($\alpha$, $\beta$, $\gamma$). we set $\alpha = 1$, $\beta = 0.3$, and $\gamma = 0.1$ in all our experiments, unless otherwise specified.

\section{Discussion on IEM}
\label{app:iem}
Recall that the IEM can be expressed as:
\begin{align}
    y_{ij}^{\gS} = f_2(f_1([y_i^{\text{agg}},y_j^{\text{agg}}])).
\end{align}
In other words, the IEM operates by accepting concatenated representations of the query and passage and determining their relevancy. Fundamentally, through $f_1$, it tries to emulate the cross-attention mechanism present in the teacher model, fostering an information exchange between the passage and query. Depending on whether the relationship between $[y_i^{\text{agg}},y_j^{\text{agg}}]$ is symmetric or asymmetric, IEM directs the input to the relevant branch of MLP $f_2 \in \{f_2^\text{sym}, f_2^\text{asym}\}$, which accordingly outputs a "yes" or "no" representation. This differentiation allows IEM to adeptly handle both types of semantic search, mirroring the diverse prompts $\mathbf P\in\{\mathbf P^\text{sym}, \mathbf P^\text{asym}\}$  utilized by the teacher model during inference.

To evaluate IEM's performance against traditional cosine similarity measures, we engaged in a comprehensive experimental suite, the details of which are provided in Appendix~\ref{app:branch}. Within this analysis, we focused on the SNLI-zh and T2Ranking datasets to represent symmetric and asymmetric semantic relationships, respectively. Models trained on these datasets include a dual-branch version (D2LLM-dual) and two specialized models (D2LLM-sym and D2LLM-asym). Additionally, we introduced cosine similarity into this comparative study, resulting in models denoted as D2LLM-cos-mixed, D2LLM-cos-sym, and D2LLM-cos-asym. The results in Table~\ref{tab:mixed_training} demonstrate that D2LLM-dual exhibits superior capability in handling both symmetric and asymmetric search tasks compared to the D2LLM-cos-mixed model. This can be attributed to the dual-branch structure of the IEM, which allows each semantic relationship type to be learned independently during training, thereby minimizing cross-interference. Conversely, cosine similarity, as implicitly mentioned in~\cite{2018Specialising} and~\cite{muennighoff2022sgpt}, struggles to distinguish between these relationship types accurately. Moreover, the enhanced performance of the D2LLM-sym and D2LLM-asym, compared to their cosine-based counterparts, further substantiates the proficiency of the learnable MLP components $f_1$ and $f_2$ within the IEM at navigating semantic nuances more adeptly than cosine similarity.

Nevertheless, it is essential to acknowledge the limitations of the IEM, in comparison with cross-encoders (i.e., the teacher model LLM-ce or LLM-ce-ft). As shown in Tables~\ref{tab:nli_results}, \ref{tab:sts_results} and \ref{tab:ir_results}, IEM does not yet match the performance of cross-encoder models, which is primarily due to its mechanism of facilitating information exchange only after the separate derivation of query and passage embeddings. This approach overlooks the nuanced relationship modeling achieved through the continuous cross-attention layers in the cross-encoders.

\section{Efficiency Enhancement in Real-world Applications}
In real-world applications, especially with larger datasets, bi-encoders, despite their capability to pre-compute vectors of passages, may face significant expenses when performing full-scale vector retrieval. To ensure efficient search, various technologies~\cite{jegou2010product,ge2013optimized} have been put forward, of which a representative solution is quantization-based Approximate Nearest Neighbor (ANN) methods. These methods compress the vector space through quantization to enhance retrieval efficiency and reduce storage demands while only marginally sacrificing accuracy, thereby markedly accelerating search speeds. These techniques are equally applicable to D2LLM. By applying quantization techniques to both the embeddings produced by the PMA and the network parameters of the IEM (which functions as the distance metric), we can slightly compromise accuracy to significantly enhance search speed. This improvement in efficiency can, in turn, contribute to better ranking performance in subsequent procedures.
\section{Bad Case Study}
\label{app:bad_case}
For the NLI task, we present two bad cases, one representing an entailment pair, and the other a contradiction pair. In addition to the labels, we have provided the student model's prediction and the teacher logits. The specific results are as follows:\\
\textbf{NLI bad case \#1:}\\
Sentence A: There is a tennis court in this city, and they are the first batch of customers. \\
Sentence B: There are customers who came to the tennis court before them.\\
Student's Prediction: 0.5959\\
Teacher's logits: 0.9141\\
Label: 0 (Contradiction)\\
\textbf{NLI bad case \#2:}\\
Sentence A: The operator said that the two parties were bargaining. \\
Sentence B: There were at least three people.\\
Student's Prediction: 0.3541\\
Teacher's logits: 0.4688\\
Label: 1 (Entailment)

Similarly for the IR task, we provide two bad cases, both of which are positive examples. We present the student's rankings of these cases and the teacher logits. The specific results are as follows:\\
\textbf{IR bad case \#1:}\\
Query: What are the four major artifacts of Asia? \\
Passage: One of the artifacts: motorcycle. Motorcycle can be said to be a status symbol in India.\\
Student's Rank: 32\\
Teacher's logits: 0.081\\
Label: 1 (Correct match)\\
\textbf{IR bad case \#2:}\\
Query: Why does my phone always show that there are messages that cannot be opened? \\
Passage: If you have a smart phone, you can try flashing it.\\
Student's Rank: 46\\
Teacher's logits: 0.135\\
Label: 1 (Correct match)

We suggest that these "bad cases" may stem from the teacher model's initial misjudgment, which inadvertently hinders the student model's learning process. To address this, we propose two potential strategies to refine the teacher model's predictions: First, for scenarios with limited resources, applying In-Context Learning~\cite{min2022rethinking} by including examples of these errors during training could enhance the teacher's ability to provide more accurate logits. Second, with sufficient resources, fine-tuning the teacher model can improve its performance while being mindful to avoid catastrophic forgetting~\cite{kirkpatrick2017overcoming} to ensure it retains its generalized learning capabilities. Both strategies signify promising areas for our future explorations.
\end{document}